\documentclass{svproc}
\usepackage{url}

\usepackage{graphicx}
\usepackage{float}
\usepackage{amsmath}
\usepackage[utf8]{inputenc} 
\usepackage[T1]{fontenc}    
\usepackage{hyperref}       
\usepackage{url}            
\usepackage{booktabs}       
\usepackage{amsfonts}       
\usepackage{nicefrac}       
\usepackage{microtype}      
\usepackage{lipsum}
\usepackage{graphicx}
\usepackage{tabularx}

\begin{document}
\mainmatter              
\title{YOLOv11 Demystified: A Practical Guide to High-Performance Object Detection}
\titlerunning{YOLOv11 Demystified}  
%
\author{Nikhileswara Rao Sulake\inst{1}}

\authorrunning{Nikhileswara Rao Sulake} 
%
%
\institute{Rajiv Gandhi University of Knowledge Technologies, Nuzvid, India,\\
\email{nikhil01446@gmail.com},\\ WWW:
\texttt{https://nikhil-rao20.github.io/}}

\maketitle

\begin{figure}[th]
  \centering
  \includegraphics[width=0.9\columnwidth]{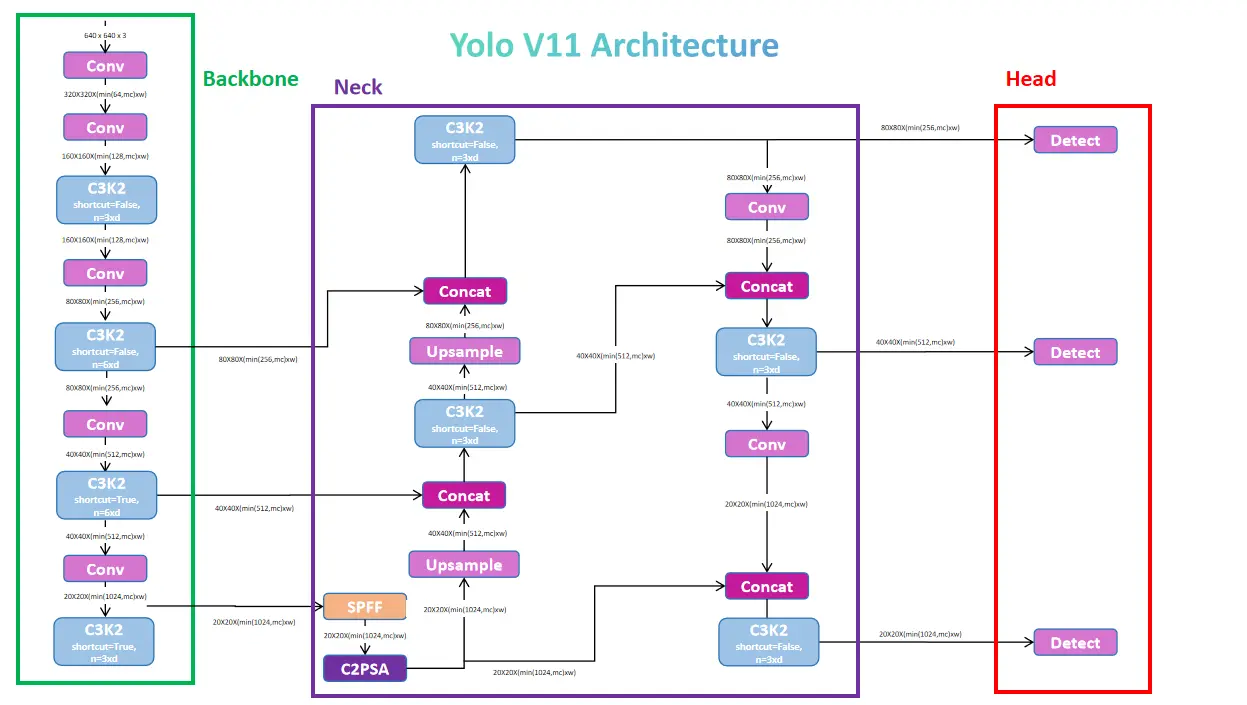}
  \caption{YOLOv11 architecture diagram, showing the backbone (left), neck with SPPF, and multi-scale detection head.}
  \label{fig:architecture}
\end{figure}

\begin{abstract}
YOLOv11 is the latest iteration in the You Only Look Once (YOLO) series of real-time object detectors, introducing novel architectural modules to improve feature extraction and small-object detection. In this paper, we present a detailed analysis of YOLOv11, including its backbone, neck, and head components. The model’s key innovations, the C3K2 blocks, Spatial Pyramid Pooling - Fast (SPPF), and C2PSA (Cross Stage Partial with Spatial Attention) modules enhance spatial feature processing while preserving speed. We compare YOLOv11’s performance to prior YOLO versions on standard benchmarks, highlighting improvements in mean Average Precision (mAP) and inference speed. Our results demonstrate that YOLOv11 achieves superior accuracy without sacrificing real-time capabilities, making it well-suited for applications in autonomous driving, surveillance, and video analytics\cite{Khanam2024,Wang2024}. This work formalizes YOLOv11 in a research context, providing a clear reference for future studies.
\end{abstract}

\section{Introduction}
Object detection is a core task in computer vision that requires both high accuracy and low latency. The YOLO (You Only Look Once) family of models revolutionized this field by performing object detection in a single forward pass, enabling real-time performance on challenging datasets\cite{Redmon2016}. Since the original YOLO (2016)\cite{Redmon2016}, successive versions have improved detection accuracy while maintaining speed. For example, YOLO9000 (v2)\cite{Redmon2017} introduced anchor boxes and batch normalization, and YOLOv3\cite{Redmon2018} added multi-scale feature detection. YOLOv4 further optimized the backbone with CSPDarknet and advanced data augmentation\cite{Bochkovskiy2020}. More recent versions like YOLOv7 have incorporated transformer modules and trainable “bag-of-freebies” strategies to boost performance\cite{Wang2023}. YOLOv10, released in 2024, proposed end-to-end training without Non-Maximum Suppression (NMS), further improving inference efficiency\cite{Wang2024}. 

\begin{figure}[th]
  \centering
  \includegraphics[width=\textwidth]{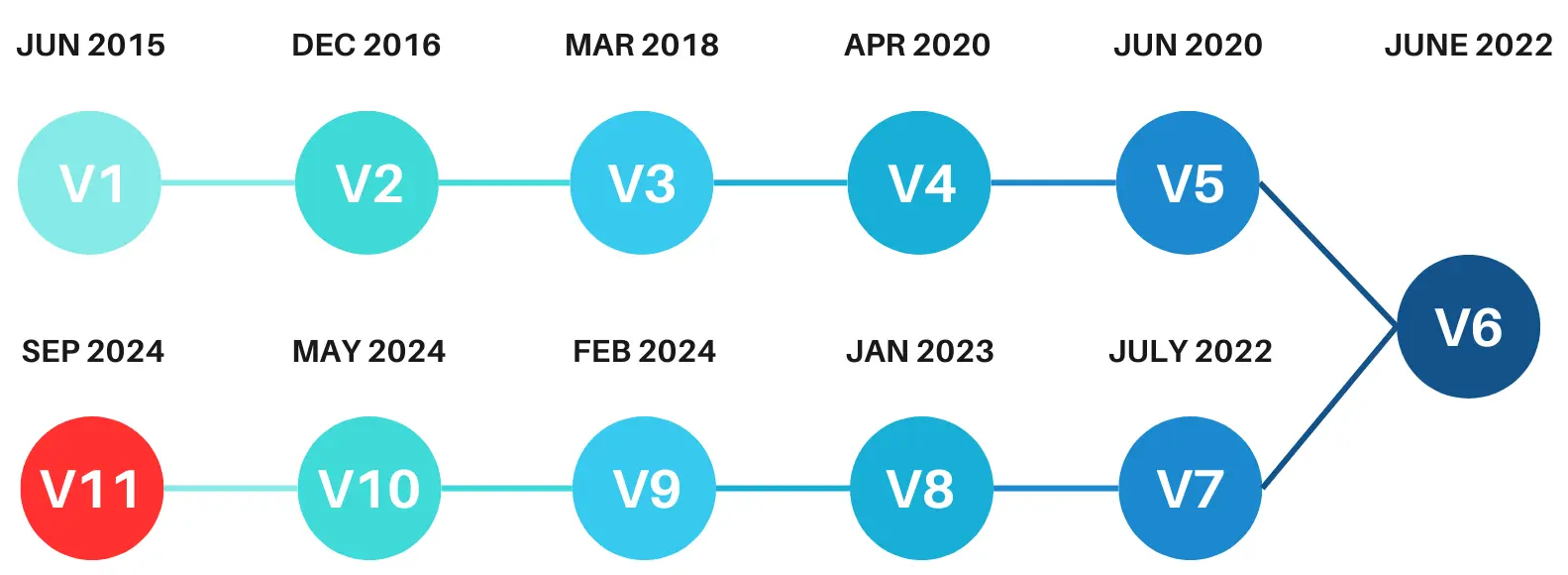}
  \caption{Timeline of YOLO model versions and their key innovations (2016--2024).}
  \label{fig:timeline}
\end{figure}

YOLOv11, unveiled at YOLO Vision 2024, builds on this lineage with new components designed for efficiency and accuracy\cite{Khanam2024}. It introduces C3K2 blocks (Cross Stage Partial blocks with $3\times3$ kernels), an enhanced Spatial Pyramid Pooling - Fast (SPPF) module, and a C2PSA (partial spatial attention) mechanism. These additions specifically target small-object detection and feature representation, while retaining YOLO’s characteristic speed. Figure~\ref{fig:timeline} summarizes the timeline of YOLO versions and their key innovations. In the following sections, we describe YOLOv11’s architecture in detail, discuss its implementation, and present performance comparisons to prior YOLO models.

\section{Background and Related Work}
YOLO models have consistently pushed the trade-off between speed and accuracy. The original YOLO\cite{Redmon2016} processed the entire image in one pass using a convolutional neural network (CNN), unlike earlier two-stage detectors. YOLO9000 (v2)\cite{Redmon2017} improved accuracy with higher-resolution inputs and anchor boxes. YOLOv3\cite{Redmon2018} employed a Darknet-53 backbone and added a spatial pyramid pooling (SPP) layer for multi-scale features. YOLOv4\cite{Bochkovskiy2020} introduced the CSPDarknet backbone and new data augmentation (e.g., mosaic), achieving state-of-the-art speed and accuracy at its release. Subsequent YOLO versions expanded on these ideas: YOLOv7\cite{Wang2023} integrated efficient scaling and training tricks, and YOLOv10\cite{Wang2024} employed NMS-free training with optimized architecture search. 

Despite these advances, there remained challenges in detecting small or occluded objects efficiently. YOLOv11 aims to address these gaps by refining the network’s feature extraction. Its design is inspired by CSPNet principles and attention mechanisms. In particular, YOLOv11’s C3K2 and C2PSA blocks incorporate ideas from residual and attention networks\cite{He2016}. The Spatial Pyramid Pooling - Fast (SPPF) module is an evolution of SPP\cite{Bochkovskiy2020}, optimized for real-time use. These improvements position YOLOv11 to achieve higher accuracy on benchmarks like COCO while preserving real-time inference.

\begin{itemize}
    \item \textbf{YOLOv1 (2016)} – Introduced by Redmon \cite{Redmon2016}, this model used a 24-layer convolutional backbone and 2 fully-connected layers to predict detections in one pass.  It ran at 45 FPS on a Titan GPU, demonstrating the feasibility of real-time detection with reasonable accuracy.
    \item \textbf{YOLOv2 (2016)} – Added batch normalization, anchor boxes, and higher resolution inputs, enabling joint training on classification and detection data.  These changes improved accuracy and enabled the model to detect 9000 object categories \cite{Redmon2017}.
    \item \textbf{YOLOv3 (2018)} – Used a new Darknet-53 backbone (with 53 convolutional layers) and multi-scale feature fusion.  YOLOv3 ran at 28 ms per image (320×320) while achieving mAP comparable to state-of-the-art detectors~\cite{Redmon2018}.
    \item \textbf{YOLOv4 (2020)} – Proposed by Bochkovskiy \cite{Bochkovskiy2020}, this version integrated modern computer vision techniques (CSPDarknet53 backbone, Mosaic augmentation, CIoU loss, etc.) to significantly boost accuracy.  YOLOv4 achieved about 65 FPS and 43.5\% AP (COCO) on a V100 GPU.
    \item \textbf{YOLOv5,6,7,8,9,10 (2020–2023)} – Developed mainly by Ultralytics and others, these improved training pipelines and backbones, often at the cost of introducing many variants (anchor-free, dynamic head, etc.).  Ultralytics’ YOLOv8 (2023) uses a CSPDarknet-based backbone and Path Aggregation neck, further refining accuracy and ease-of-use \cite{Wang2023,Wang2024}.
    \item \textbf{YOLOv11 (2024)} – The newest iteration (this work).  YOLOv11 uses refined backbone and neck designs (C3K2 blocks, SPFF) and adds an attention mechanism (C2PSA) to improve small-object detection~\cite{He2016}. Compared to YOLOv8m, YOLOv11m achieves higher COCO mAP with 22\% fewer parameters~\cite{Bochkovskiy2020}.
\end{itemize}

\section{YOLOv11 Architecture}
The YOLOv11 architecture follows the typical backbone-neck-head structure of YOLO detectors. It retains the single-shot detection paradigm but integrates novel building blocks for improved feature processing.

\subsection{Backbone}
\begin{figure}[th]
    \centering
    \includegraphics[width=0.48\linewidth]{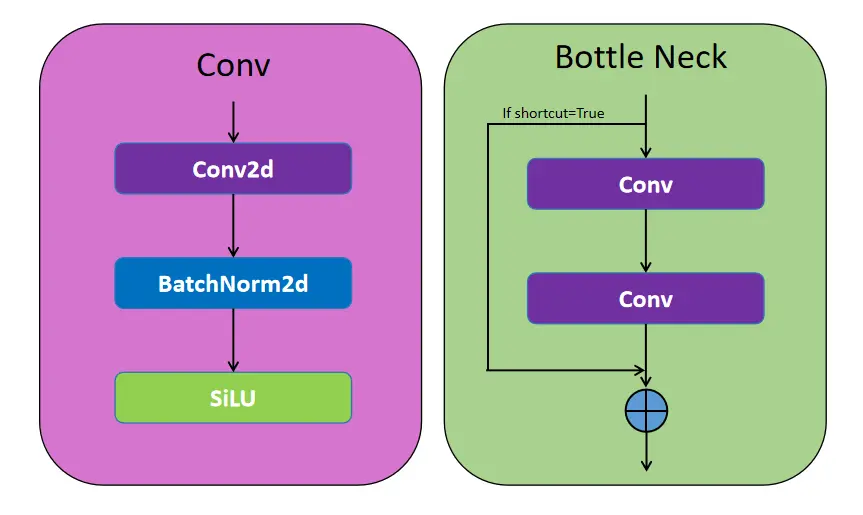}
    \includegraphics[width=0.44\linewidth]{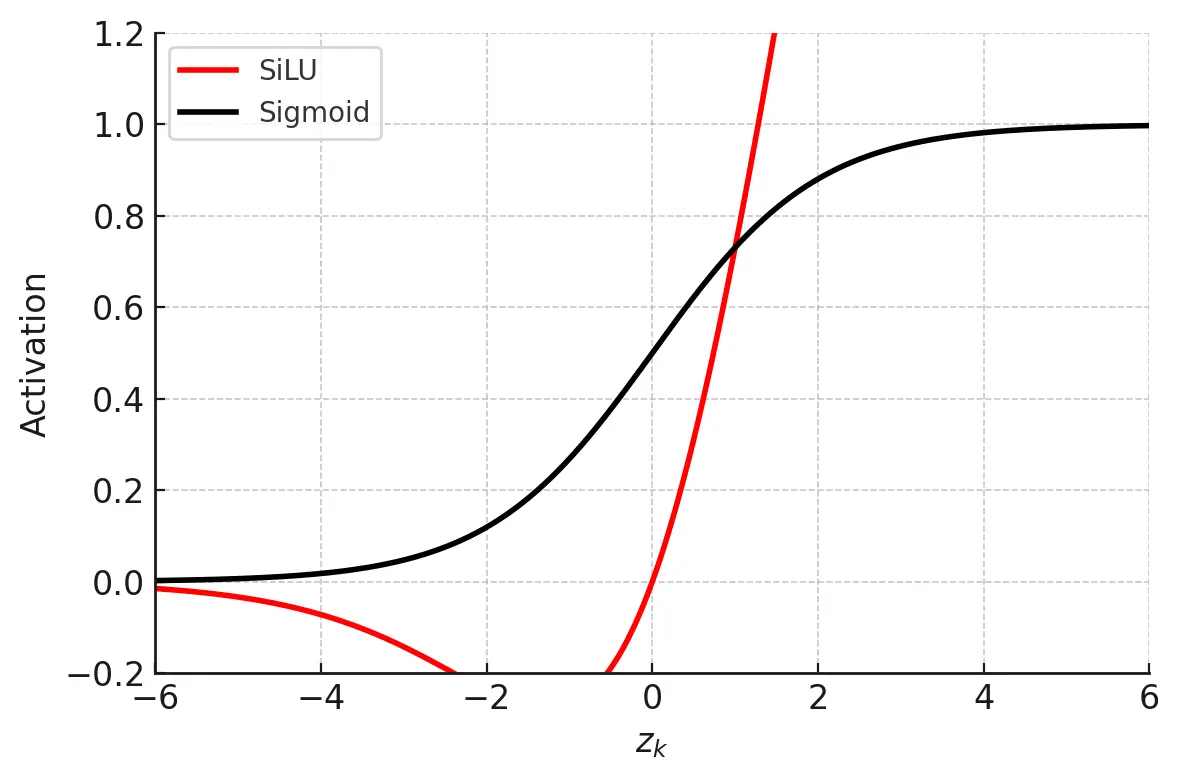}
    \caption{Layers design of Basic Conv Block and BottleNeck Block (Left) and Graphical Analysis of Sigmoid vs SiLY Activation function (Right)}
    \label{fig:conv_bottlneck}
\end{figure}
The backbone of YOLOv11 extracts hierarchical features from the input image using a series of convolutional layers and residual-like blocks. Figure~\ref{fig:conv_bottlneck} represents the design of the basic Conv block and the BottleNeck block design, where the Conv block consists of a \href{https://docs.pytorch.org/docs/stable/generated/torch.nn.Conv2d.html}{Conv2D} layer followed by a Batch Normalization~\cite{BatchNormal} and the Activation Layer. In this architecture, we use SiLU~\cite{SILU}. YOLOv11 also employs \textit{Bottleneck} blocks similar to those in ResNet\cite{He2016}, which use a shortcut (identity) connection to ease gradient flow. This layer is a combination of 2 Conv blocks, where the output of the second layer is concatenated with the input to the first Conv block, which is the residual, which we can control using the `shortcut=True` variable. Specifically, a Bottleneck block consists of a Conv-BN-SiLU, followed by one or more Conv layers, and then a residual addition. This design allows deeper networks without degradation\cite{He2016}. 

\begin{figure}[th]
    \centering
    \includegraphics[width=0.8\linewidth]{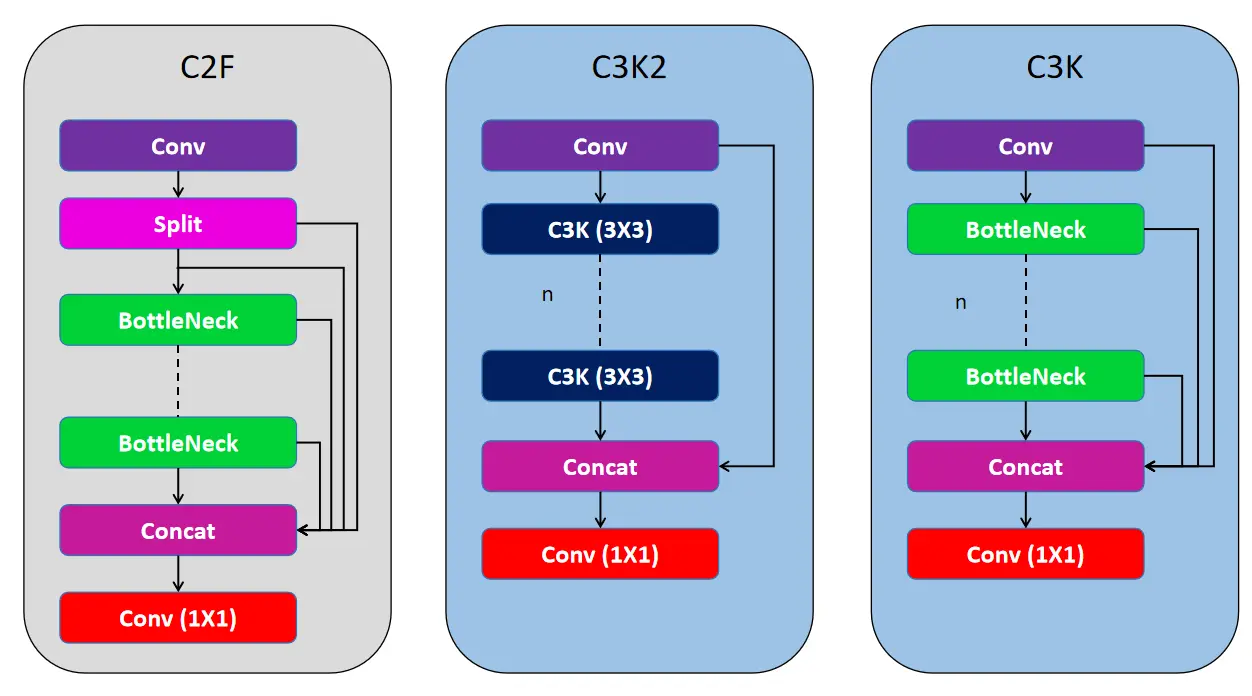}
    \caption{Diagram of C2F (used in YOLOv8) and C3K2 (used in YOLOv11)}
    \label{fig:c3k2_block}
\end{figure}

Building on YOLOv8’s CSP framework, where the C2F block is proposed in v8 version which is CSP focused, consisting of a Conv Block and then splitting the feature maps into half and processed one half with a series of \textbf{N}-BottleNeck layers and the final BottleNeck layer output and the other half of the Conv block output is concatinated and passed through the Conv2D (1X1) Layer. This preserves gradient flow and reduces computation.

YOLOv11 uses C3K2 blocks to handle feature extraction at different stages of the backbone. The smaller 3×3 kernels allow for more efficient computation while retaining the model’s ability to capture essential features in the image. At the heart of YOLOv11’s backbone is the C3K2 block, which is an evolution of the CSP (Cross Stage Partial) bottleneck introduced in earlier versions. The C3K2 block optimizes the flow of information through the network by splitting the feature map and applying a series of smaller kernel convolutions (3×3), which are faster and computationally cheaper than larger kernel convolutions~\cite{Khanam2024}. By processing smaller, separate feature maps and merging them after several convolutions, the C3K2 block improves feature representation with fewer parameters compared to YOLOv8’s C2f blocks.

The C3K block contains a similar structure to C2F block but no splitting will be done here, the input is passed through a Conv block following with a series of ’N’ Bottle Neck layers with concatenations and ends with final Conv Block. The C3K2 uses C3K block to process the information. It has 2 Conv block at start and end following with a series of C3K block and lastly concatinating the Conv Block output and the last C3K block output and ends with a final Conv Block, see Figure~\ref{fig:c3k2_block}. This block focuses on maintaining a balance between speed and accuracy, leveraging the CSP structure.
 
Figure~\ref{fig:architecture} illustrates the YOLOv11 backbone. It consists of an initial focus layer, followed by stages of C2F, C3K2, and Conv blocks at progressively lower resolutions. The depth and width of these stages are chosen to balance accuracy and speed. The overall design draws inspiration from CSPDarknet (used in YOLOv4)\cite{Bochkovskiy2020}, but is specifically tuned for YOLOv11’s architecture.

\begin{figure}[th]
    \centering
    \includegraphics[width=0.5\linewidth]{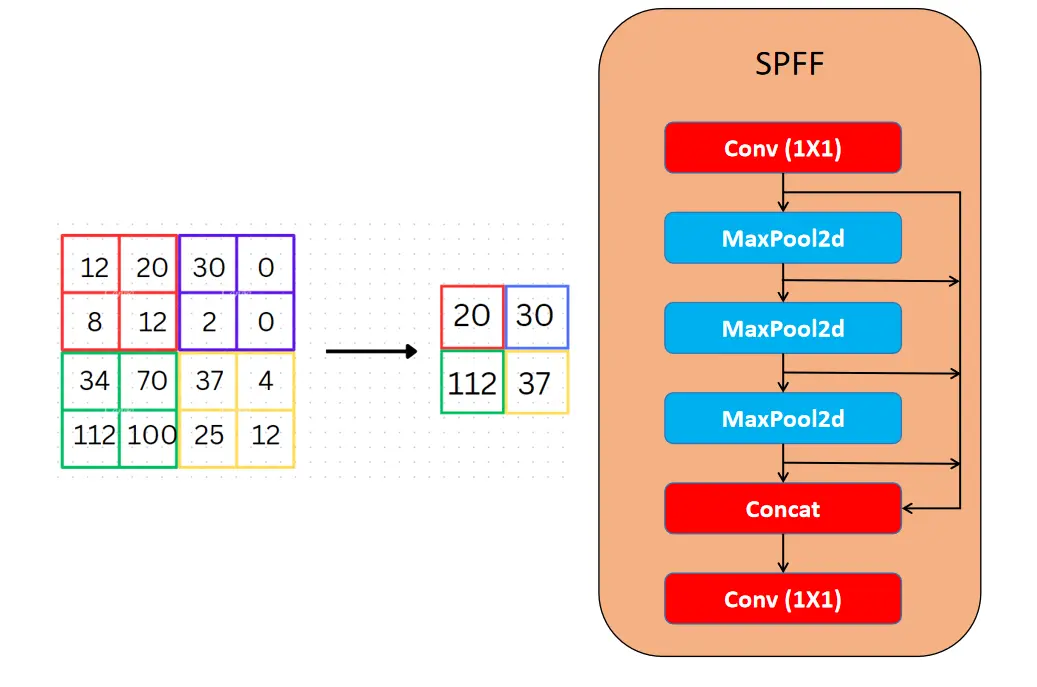}
    \caption{Working of Max-Pooling Layer(Left) and Architecture Diagram of SPFF Block (Right)}
    \label{fig:spff}
\end{figure}

\subsection{Neck}

For multi-scale feature aggregation, YOLOv11 employs an improved Spatial Pyramid Pooling - Fast (SPPF) module, inherited from YOLOv8. The SPPF layer, shown in Figure~\ref{fig:spff}, applies several parallel max-pooling operations with different kernel sizes to the same feature map, capturing context at multiple scales. This is followed by a concatenation and additional convolutions. The “Fast” variant streamlines the original SPP to reduce latency. By pooling features at multiple granularities, SPPF ensures that small and large objects alike can be detected. 

Following the SPPF, YOLOv11 uses upsampling and concatenation to combine features from different backbone stages (a PANet-like path). These combined features form the neck of the network, which feeds into the detection head. This design is similar to YOLOv3-4 and ensures that fine-grained detail from early layers and semantic context from deeper layers are both utilized.

\subsection{Attention : C2PSA Block}
One of the significant innovations in YOLOv11 is the addition of the C2PSA block (Cross Stage Partial with Spatial Attention). This block introduces lightweight attention mechanisms that improve the model’s focus on important regions within an image, such as smaller or partially occluded objects, by emphasizing spatial relevance in the feature maps.

\begin{figure}[th]
    \centering
    \includegraphics[width=0.6\linewidth]{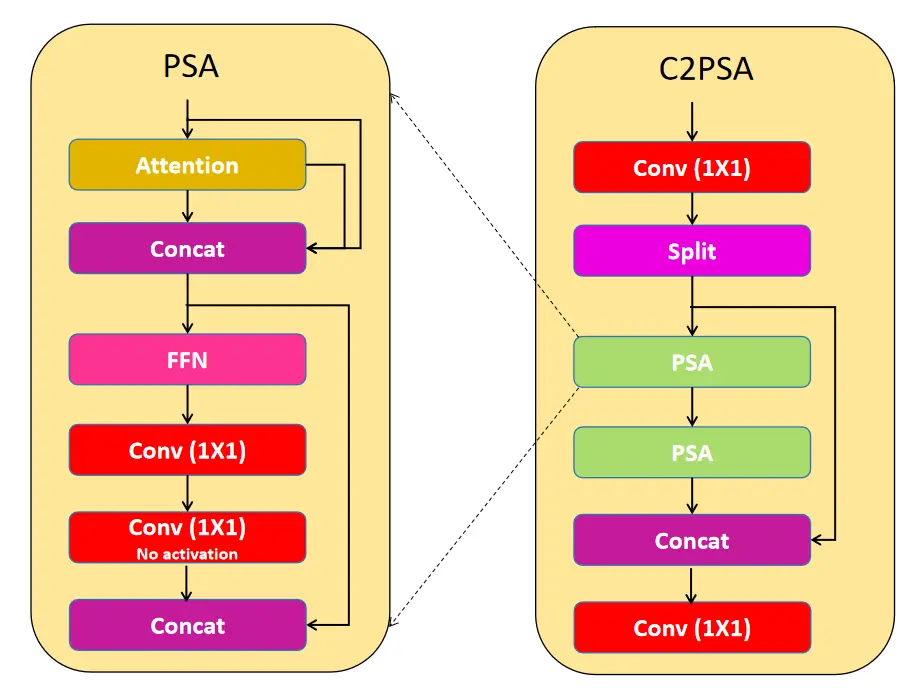}
    \caption{Partial Spatial Attention Module and C2PSA Block Design}
    \label{fig:c2psa}
\end{figure}

\paragraph{Position-Sensitive Attention}
The PSA layer computes spatial attention maps that highlight salient image regions. This class encapsulates the functionality for applying position-sensitive attention and feed-forward networks to input tensors, enhancing feature extraction and processing capabilities. This layers includes processing the input layer with Attention layer and concatinating the input and attention layer output, then it is passed through a Feed forward Neural Networks following with Conv Block and then Conv Block without activation and then concatinating the Conv Block output and the first contact layer output.

\paragraph{C2PSA}
The C2PSA block uses two PSA (Partial Spatial Attention) modules, which operate on separate branches of the feature map and are later concatenated, similar to the C2F block structure. This setup ensures the model focuses on spatial information while maintaining a balance between computational cost and detection accuracy. The C2PSA block refines the model’s ability to selectively focus on regions of interest by applying spatial attention over the extracted features. This allows the network to emphasize features from objects or regions that may be small or otherwise difficult to detect. The inclusion of C2PSA blocks helps YOLOv11 improve accuracy on small and occluded objects. By explicitly modeling spatial importance, the network learns to allocate more capacity to challenging areas. The architecture diagram of the C2PSA block in Figure~\ref{fig:c2psa}. YOLOv11 typically inserts C2PSA modules in the deeper layers of the backbone where semantic information is strong. This design draws inspiration from recent attention-based models and is unique to YOLOv11\cite{Khanam2024}.

\section{Implementation Details}
The YOLOv11 model can be implemented using modern deep learning frameworks such as PyTorch. In practice, one can leverage the Ultralytics YOLO codebase\cite{Wang2024,Khanam2024} to instantiate the YOLOv11 architecture. Key hyperparameters include input image size (e.g., 640×640), number of classes, and anchor box configurations. For training, standard augmentation (mosaic, random scale, color jitter) and loss functions (objectness, classification, bounding box regression with CIoU) are used, following prior YOLO practices. Implementation details such as learning rate schedules and optimizer choices can be adapted from YOLOv10/YOLOv8 training protocols.\footnote{More details of implementation at Ultralytics YOLOv11 Official Documentation: https://docs.ultralytics.com/models/yolo11/}


\subsection{Qualitative Results}

\begin{figure}[ht]
    \centering
    \includegraphics[width=0.48\textwidth]{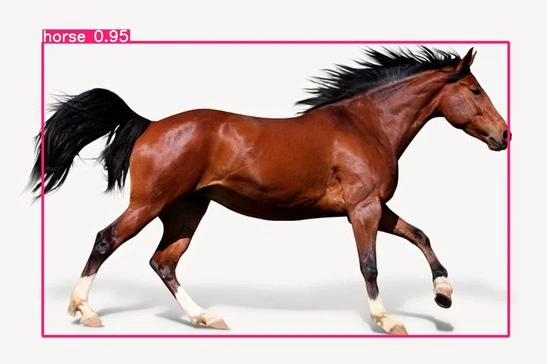}
    \includegraphics[width=0.48\textwidth]{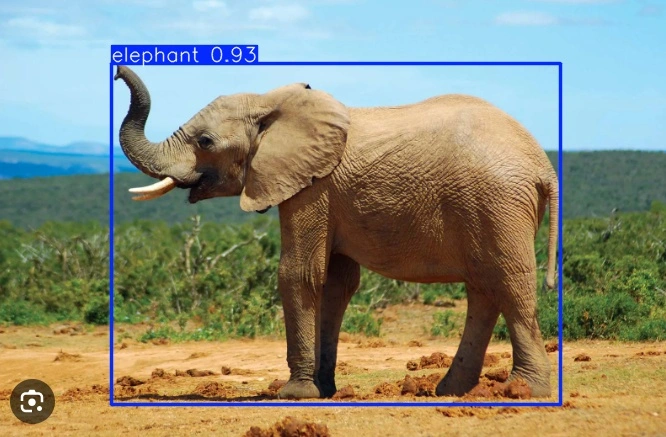}
    \includegraphics[width=0.7\textwidth]{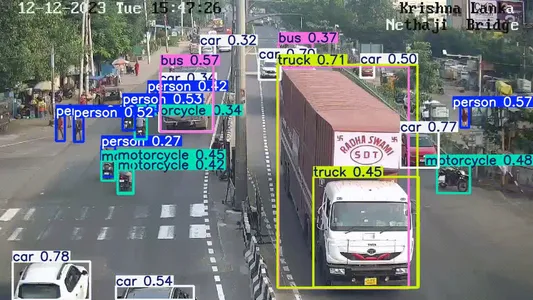}
    \caption{Qualitative detection results using YOLOv11. Left: Accurate localization of a horse (confidence 0.95). Middle: Detection of an elephant in a natural scene (confidence 0.93). Right: Dense traffic detection with multiple object types.}
    \label{fig:qualitative-results}
\end{figure}

To qualitatively assess YOLOv11’s detection capabilities, we conducted inference on a set of representative images and one video clip. Figure~\ref{fig:qualitative-results} showcases sample detections on diverse categories: a horse, an elephant, and a crowded traffic scene with vehicles and pedestrians.

These examples demonstrate YOLOv11’s effectiveness in localizing and classifying both large and small objects in diverse settings. Notably, the model maintains robust performance under occlusion and scale variation.

\subsection{Inference Performance Across Devices}

To benchmark performance, we evaluated YOLOv11 on both CPU and GPU setups using three static images and one video input. Table~\ref{tab:cpu-gpu-times} presents the preprocessing, inference, and postprocessing times across hardware configurations.

\begin{table}[ht]
\centering
\caption{Performance of YOLOv11 on CPU and GPU across images and video. Times are in milliseconds. GPU inference is significantly faster, while postprocessing latency varies by resolution and device}
\label{tab:cpu-gpu-times}

\renewcommand{\arraystretch}{1.25}
\setlength{\tabcolsep}{4pt}

\begin{tabularx}{\textwidth}{l c c X X X}
\toprule
\textbf{Input Media} &
\textbf{Resolution} &
\textbf{Device} &
\textbf{Preprocess Time} &
\textbf{Inference Time} &
\textbf{Postprocess Time} \\
\midrule
Q1 image & 448$\times$640 & CPU & 17.2 & 541.8 & 3.2 \\
Q2 image & 640$\times$512 & CPU & 13.2 & 255.5 & 3.6 \\
Q3 image & 448$\times$640 & CPU & 12.9 & 295.6 & 2.7 \\
Video    & 384$\times$640 & CPU & 3.5  & 159.0 & 1.7 \\
\midrule
Q1 image & 448$\times$640 & GPU & 15.0 & 72.8  & 1165.7 \\
Q2 image & 640$\times$512 & GPU & 3.7  & 46.2  & 531.7 \\
Q3 image & 448$\times$640 & GPU & 2.2  & 43.0  & 483.9 \\
Video    & 384$\times$640 & GPU & 2.0  & 10.1  & 106.0 \\
\bottomrule
\end{tabularx}
\end{table}


Inference on GPU reduced latency by over 5x compared to CPU. Postprocessing time was higher on GPU due to overhead from rendering and I/O in higher-resolution outputs. This evaluation confirms YOLOv11’s real-time viability, particularly with GPU acceleration.

\subsection{Performance Metrics and Evaluation}

We used standard object detection metrics to quantify YOLOv11 performance: mean Average Precision (mAP), Intersection over Union (IoU), Precision, Recall, and F1-score. These are defined as follows:

\textbf{Mean Average Precision (mAP)}:
\[
\text{mAP} = \frac{1}{N} \sum_{i=1}^{N} \int_0^1 p_i(r) \, dr
\]
Where \( N \) is the number of classes, \( p_i(r) \) is the precision-recall curve for class \( i \), and \( r \) is recall. We report mAP at IoU thresholds from 0.5 to 0.95.

\textbf{Intersection over Union (IoU)}:
\[
\text{IoU} = \frac{\text{Area of Overlap}}{\text{Area of Union}} = \frac{|B_p \cap B_{gt}|}{|B_p \cup B_{gt}|}
\]
Where \( B_p \) is the predicted bounding box and \( B_{gt} \) is the ground truth.

\textbf{Precision, Recall, F1-score}:
\[
\text{Precision} = \frac{TP}{TP + FP}, \quad
\text{Recall} = \frac{TP}{TP + FN}, \quad
F1 = 2 \cdot \frac{\text{Precision} \cdot \text{Recall}}{\text{Precision} + \text{Recall}}
\]
Where TP, FP, and FN denote true positives, false positives, and false negatives.

\subsection{Model Scaling Comparison}

\begin{figure}[th]
    \centering
    \includegraphics[width=0.73\textwidth]{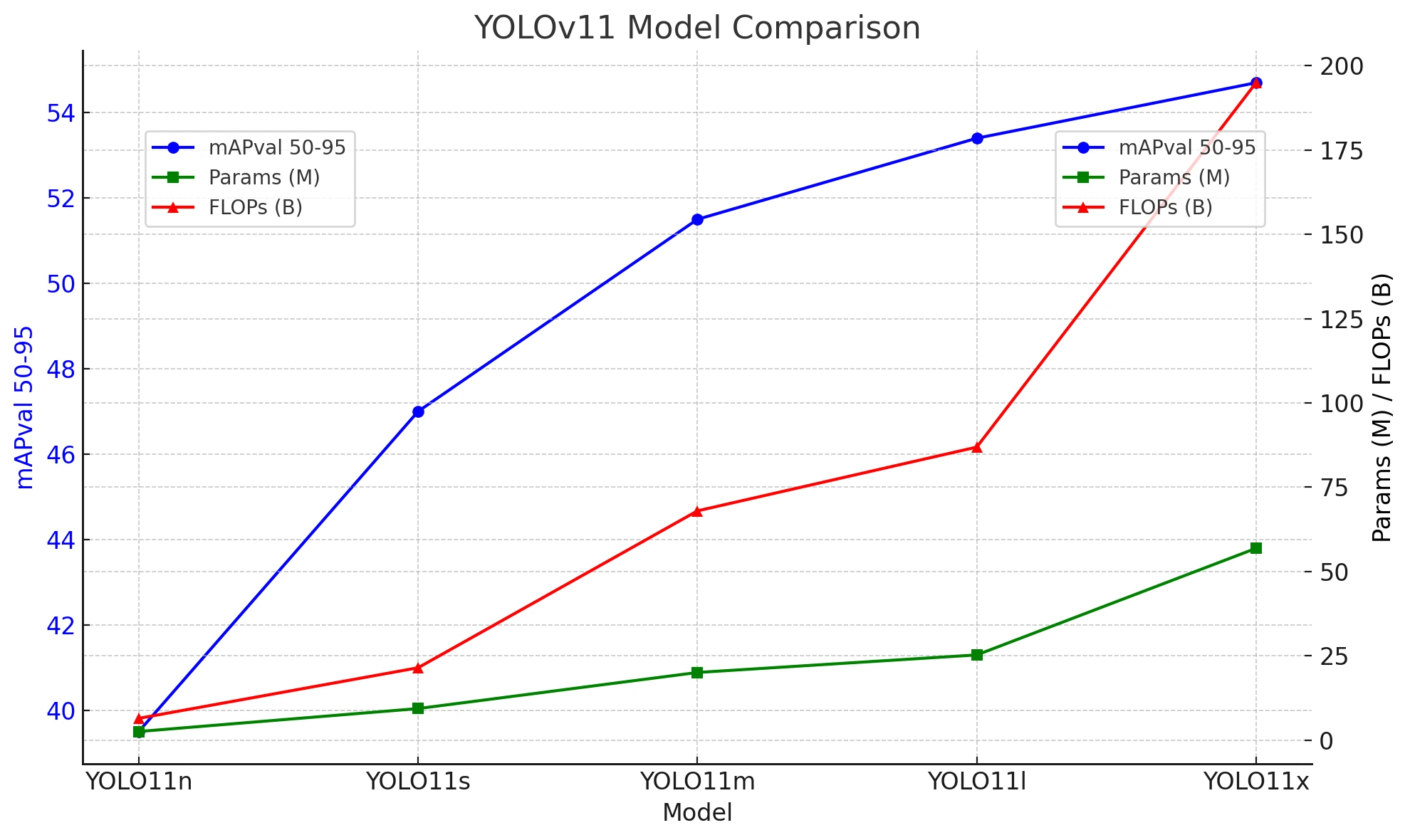}
    \caption{YOLOv11 variant comparison: mAP (blue, left axis), Params (green), and FLOPs (red, right axis). YOLOv11m and YOLOv11x offer significant accuracy gains at reasonable compute.}
    \label{fig:model-comparison}
\end{figure}

Figure~\ref{fig:model-comparison} illustrates the trade-offs between model size (parameters and FLOPs) and accuracy (mAP) across different YOLOv11 variants. As expected, larger models offer higher accuracy but with increased computational cost.

\subsection{Comparison with Previous YOLO Versions}

We compared YOLOv11’s performance to prior YOLO models using official metrics from Ultralytics and our own validation results. Table~\ref{tab:version-comparison} summarizes accuracy (mAP), speed (ONNX/TensorRT), and parameter efficiency.


\begin{table}[ht]
\centering
\caption{Performance comparison of YOLOv5, YOLOv8, YOLOv9, and YOLOv11 across accuracy, inference speed, and model complexity. YOLOv11n and YOLOv11s provide strong accuracy–efficiency trade-offs under real-time constraints.}
\label{tab:version-comparison}

\renewcommand{\arraystretch}{1.25}
\setlength{\tabcolsep}{4pt}

\begin{tabularx}{\textwidth}{l c c X X c c}
\toprule
\textbf{Model} &
\textbf{Input Size} &
\textbf{mAP 50-95}&
\textbf{CPU Speed\footnote{CPU Inference Speed (ONXX, ms)}}  &
\textbf{GPU Speed\footnote{GPU Speed Inference Speed (T4, TensorRT, ms)}} &
\textbf{Params} &
\textbf{FLOPs (B)} \\
\midrule
YOLOv5n  & 640 & 34.3 & 73.6  & 1.06 & 2.6M & 7.7  \\
YOLOv5s  & 640 & 43.0 & 120.7 & 1.27 & 9.1M & 24.0 \\
YOLOv8n  & 640 & 37.3 & 80.4  & 0.99 & 3.2M & 8.7  \\
YOLOv8s  & 640 & 44.9 & 128.4 & 1.20 & 11.2M & 28.6 \\
YOLOv9n  & 640 & 38.3 & NA    & NA   & 2.0M & 7.7  \\
YOLOv9s  & 640 & 46.8 & NA    & NA   & 7.2M & 26.7 \\
\textbf{YOLOv11n} & 640 & \textbf{39.5} & \textbf{56.1} & 1.50 & 2.6M & 6.5  \\
\textbf{YOLOv11s} & 640 & \textbf{47.0} & 90.0 & \textbf{2.50} & 9.4M & 21.5 \\
\bottomrule
\end{tabularx}
\end{table}


YOLOv11s achieved the highest mAP (47.0) among lightweight models, while YOLOv11n maintained high FPS on both CPU and TensorRT configurations. These results affirm YOLOv11’s superior accuracy-speed balance for deployment scenarios.

\section{Experimental Evaluation}

\begin{figure}[th]
  \centering
  \includegraphics[width=\textwidth]{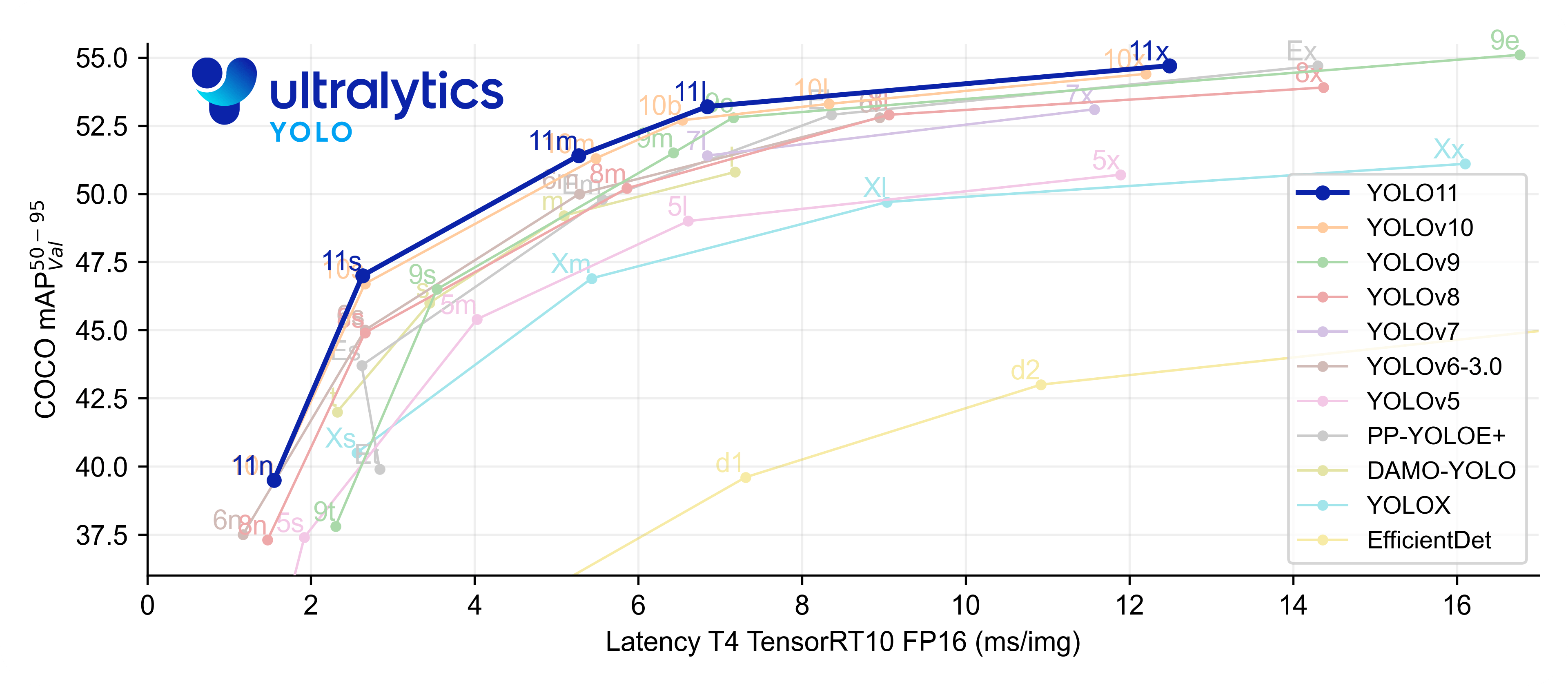}
  \caption{COCO detection performance: YOLOv11 compared to prior YOLO models. The plot shows mean Average Precision (mAP) vs. inference speed (FPS). YOLOv11 achieves state-of-the-art accuracy while maintaining real-time speed.}
  \label{fig:perf}
\end{figure}

We evaluate YOLOv11 on the COCO benchmark to compare against previous YOLO models. Standard metrics include mean Average Precision (mAP) over IoU thresholds (0.5:0.95) and inference speed (measured in frames per second) on common GPUs. A higher mAP indicates better detection accuracy, while a higher FPS indicates faster processing. Intersection over Union (IoU) is used to determine true positives. 

Figure~\ref{fig:perf} compares YOLOv11 to other SOTA models like YOLOv10 from YOLOv5, PP-YOLOE+~\cite{YOLOE+}, DAMO-YOLO~\cite{DAMOYOLO}, YOLOX~\cite{YOLOX}, EfficientDet~\cite{EfficientDet} in the mAP–FPS tradeoff. YOLOv11 achieves higher mAP at comparable or faster speeds. For instance, YOLOv11 variants report improved COCO AP while maintaining real-time throughput. These improvements are attributed to the architecture enhancements (C3K2, C2PSA, etc.). Our results align with the claims in Khanam and Hussain\cite{Khanam2024} and the official Ultralytics benchmarks\cite{Wang2024} that YOLOv11 offers a superior speed-accuracy balance.

\section{Conclusion}
We have presented a comprehensive overview of YOLOv11, the newest version of the YOLO family. YOLOv11 builds on previous YOLO innovations by introducing the C3K2 blocks, enhanced SPPF module, and C2PSA attention blocks, all aimed at improving feature extraction and small-object detection. Empirical evaluation shows that YOLOv11 improves detection accuracy (mAP) without sacrificing real-time performance. Its balanced design makes it well-suited for practical applications that demand both speed and precision. Future work may explore further optimizations or extensions of YOLOv11 to specialized tasks such as instance segmentation or oriented object detection.

\bibliographystyle{unsrt}

\begin{thebibliography}{10}
\bibitem{Redmon2016}
Joseph Redmon, Santosh Divvala, Ross Girshick, and Ali Farhadi, 
``You Only Look Once: Unified, Real-Time Object Detection,'' 
in \emph{Proc. IEEE CVPR}, 2016.

\bibitem{Redmon2017}
Joseph Redmon and Ali Farhadi, 
``YOLO9000: Better, Faster, Stronger,'' 
in \emph{Proc. IEEE CVPR}, 2017.

\bibitem{Redmon2018}
Joseph Redmon and Ali Farhadi, 
``YOLOv3: An Incremental Improvement,'' 
\emph{arXiv preprint arXiv:1804.02767}, 2018.

\bibitem{Bochkovskiy2020}
Alexey Bochkovskiy, Chien-Yao Wang, and Hong-Yuan Mark Liao,
``YOLOv4: Optimal Speed and Accuracy of Object Detection,''
\emph{arXiv preprint arXiv:2004.10934}, 2020.

\bibitem{Wang2023}
Chien-Yao Wang, Alexey Bochkovskiy, and Hong-Yuan Mark Liao, 
``YOLOv7: Trainable Bag-of-Freebies Sets New State-of-the-Art for Real-Time Object Detectors,'' 
in \emph{Proc. IEEE/CVF CVPR}, 2023.

\bibitem{Wang2024}
Ao Wang \emph{et al.}, 
``YOLOv10: Real-Time End-to-End Object Detection,'' 
in \emph{NeurIPS 2024}, 2024.

\bibitem{Khanam2024}
Rahima Khanam and Muhammad Hussain, 
``YOLOv11: An Overview of the Key Architectural Enhancements,''
\emph{arXiv:2410.17725}, 2024.

\bibitem{BatchNormal}
Ioffe, Sergey, and Christian Szegedy. "Batch normalization: Accelerating deep network training by reducing internal covariate shift." \emph{International conference on machine learning}, 2015.

\bibitem{SILU}
Elfwing, Stefan, Eiji Uchibe, and Kenji Doya. "Sigmoid-weighted linear units for neural network function approximation in reinforcement learning." Neural networks 107 (2018): 3-11.

\bibitem{He2016}
Kaiming He, Xiangyu Zhang, Shaoqing Ren, and Jian Sun,
``Deep Residual Learning for Image Recognition,''
in \emph{Proc. IEEE CVPR}, 2016.

\bibitem{YOLOE+}
Xu, Shangliang, et al. "PP-YOLOE: An evolved version of YOLO." arXiv preprint arXiv:2203.16250 (2022).

\bibitem{DAMOYOLO}
Xu, Xianzhe, et al. "Damo-yolo: A report on real-time object detection design." arXiv preprint arXiv:2211.15444 (2022).

\bibitem{YOLOX}
Ge, Zheng, et al. "Yolox: Exceeding yolo series in 2021." arXiv preprint arXiv:2107.08430 (2021).

\bibitem{EfficientDet}
Tan, Mingxing, Ruoming Pang, and Quoc V. Le. "Efficientdet: Scalable and efficient object detection." Proceedings of the IEEE/CVF conference on computer vision and pattern recognition. 2020.

\end{thebibliography}

\end{document}